\documentclass[a4paper, 10pt, conference]{ieeeconf}      

\usepackage{xcolor}
\usepackage{gensymb}
\usepackage{cite}

\usepackage{graphicx} 

\title{\LARGE \bf
GelSight360: An Omnidirectional Camera-Based Tactile Sensor for Dexterous Robotic Manipulation
}

\author{
    \authorblockN{Megha H. Tippur and Edward H. Adelson} 
        \authorblockA{Massachusetts Institute of Technology\\
    {\tt\small mhtippur@csail.mit.edu, adelson@csail.mit.edu} 
    } }

\begin{document}

\maketitle
\thispagestyle{empty}
\pagestyle{empty}

\begin{abstract}


Camera-based tactile sensors have shown great promise in enhancing a robot's ability to perform a variety of dexterous manipulation tasks. Advantages of their use can be attributed to the high resolution tactile data and 3D depth map reconstructions they can provide. Unfortunately, many of these tactile sensors use either a flat sensing surface, sense on only one side of the sensor's body, or have a bulky form-factor, making it difficult to integrate the sensors with a variety of robotic grippers. Of the camera-based sensors that do have all-around, curved sensing surfaces, many cannot provide 3D depth maps; those that do often require optical designs specified to a particular sensor geometry. In this work, we introduce GelSight360, a fingertip-like, omnidirectional, camera-based tactile sensor capable of producing depth maps of objects deforming the sensor's surface. In addition, we introduce a novel cross-LED lighting scheme that can be implemented in different all-around sensor geometries and sizes, allowing the sensor to easily be reconfigured and attached to different grippers of varying DOFs. With this work, we enable roboticists to quickly and easily customize high resolution tactile sensors to fit their robotic system's needs.

\end{abstract}

\section{INTRODUCTION}
As the use of robots in both industry and our everyday lives becomes more ubiquitous, improvements in a robot’s ability to safely and reliably complete a wide range of dexterous manipulation tasks become necessary. One way to enhance robots is to provide them with the sense of touch. When interacting with the environment, occlusions to the vision system by other objects or the manipulator itself can occur. Sometimes, no visual feedback may even be available, such as when searching in a bag or on the top shelf of a cabinet. In such cases, using tactile feedback can greatly benefit the system. 

Specifically, camera-based tactile sensors, like GelSight, have been used to successfully complete a variety of tasks \cite{wang2021gelsight, alspach2019soft,she2021cable}. Their ability to provide high resolution information and depth reconstructions can help a robot estimate the pose of an object in contact with the end effector \cite{she2021cable, liu2022gelsight} or quickly sense unforeseen collisions and prevent excessive force before an object is damaged \cite{yuan2016estimating}. 

\begin{figure}[!h]
    \centering
    \includegraphics[width=\linewidth]{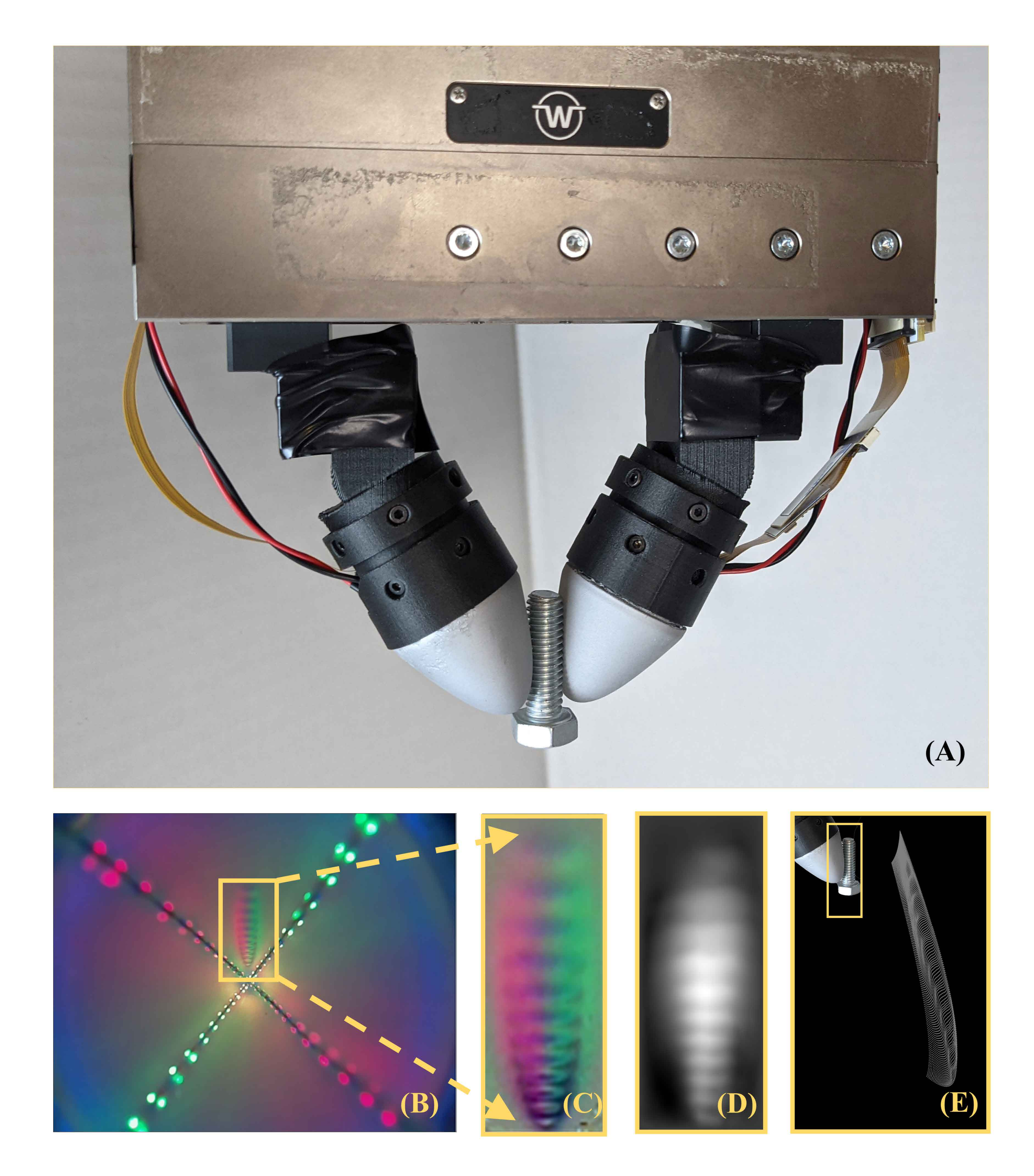}
    \vspace*{-10mm}
    \caption{\footnotesize{\textbf{(A)} Two GelSight360 sensors mounted on the WSG 50-110 (Weiss) robotic gripper with an M6 screw in its grasp. \textbf{(B)} Example of the tactile image captured by the camera housed inside the sensor's body showing the impression of the screw threads. \textbf{(C)} Zoomed in difference image of the contact region of interest (ROI). \textbf{(D)} Reconstructed depth map of the contact region. \textbf{(E)} Point cloud of the contact ROI projected onto the sensor body's surface showing the deformation of the soft gel elastomer caused by the screw threads.}}
    \label{fig:teaser}
\end{figure}
Unfortunately, many of the sensors capable of producing high resolution tactile data and depth reconstructions have a flat sensing surface and bulky form-factor, so integrating tactile sensors with robotic hands that use multiple fingers and higher DOFs can be difficult. Previous attempts at curved camera-based tactile sensors required extensive testing to ensure good illumination throughout the sensor, so alterations to the sensor's size and geometry would require reconfiguration and tuning of the design process. 
We introduce a curved, all-around, camera-based tactile sensor in a fingertip-like form factor which is capable of producing high resolution tactile data and depth reconstructions, as shown in Figure \ref{fig:teaser}. Additionally, the lighting scheme and calibration procedure introduced in this work can be generalized to a variety of different sensor shapes and sizes, making it easier and more accessible for roboticists to incorporate tactile sensing into their robotic systems. The all-around sensing nature of this design also allows for the sensors to be employed on a variety of grippers, both anthropomorphic and non-anthropomorphic, with varying DOFs.  

\section{RELATED WORKS}
In recent years, a great deal of progress has been made in the development of tactile sensing technologies for integration with robotic grippers for dexterous manipulation tasks.
\cite{kappassov2015tactile}. A variety of transduction methods, ranging from capacitive \cite{zainuddin2015resistive, muhlbacher2015responsive, heyneman2012biologically}, resistive \cite{sundaram2019learning, fishel2012sensing, ntagios2020robotic}, piezoelectric \cite{zhang2022finger, yong2022soft, tang2019design}, magnetic \cite{bhirangi2021reskin}, and vision-based \cite{alspach2019soft, ward2018tactip, lambeta2020digit, shimonomura2019tactile, yamaguchi2016combining} have been used to help robots better interact with their environments by providing them with the sense of touch. In particular, visuotactile sensors have gained popularity in robotics due to their ability to provide high resolution information over a variety of sensing modalities, such as contact localization \cite{yamaguchi2016combining, li2014localization}, normal force estimation \cite{sun2022soft, taylor2022gelslim, yuan2017gelsight}, slip detection \cite{yuan2015measurement, li2018slip}, and object geometry estimation \cite{alspach2019soft,wang2021gelsight, 
yuan2017gelsight, do2022densetact, yin2022multimodal}. 

\subsection{GelSight and Related Sensors}

Our sensors use similar lighting principles, designs, and stereo algorithms introduced in prior GelSight works \cite{yuan2017gelsight} to obtain 3D geometric reconstructions of objects deforming the surface of the sensor, but now, in a compact, curved form-factor with tactile sensing capabilities all around the surface of the sensor. The basis of GelSight and related sensors use a camera embedded behind a piece of clear acrylic bonded to a soft, gel elastomer. The elastomer is painted with a reflective coating to capture any deformations on the surface of the gel.
Additionally, by illuminating the sensor with different colored LEDs (red, green, and blue), photometric stereo methods can be used to estimate the surface normals of the contact region \cite{johnson2009retrographic, yuan2017gelsight}. These normals can then be used to calculate the depth maps of the sensor’s surface \cite{yuan2017gelsight, wang2021gelsight}. 3D depth reconstructions have shown to be extremely useful in a large number of dexterous robotic manipulations tasks, such as object shape perception \cite{li2014localization}, texture and hardness classification \cite{li2013sensing, yuan2016estimating, yuan2018active}, pose estimation \cite{wang2021gelsight, bauza2019tactile}, and deformable object manipulation \cite{she2021cable, huang2022understanding}. However, even though these sensors are able to capture tactile data with a high spatial resolution, one drawback is that many GelSight (and related) sensors have a flat sensing surface. 

GelSight sensors with more unconventional and non-flat sensing surfaces have also been developed \cite{romero2020soft, patel2020digger, liu2022gelsight}. Romero and Adelson were able to use the principles of photometric stereo to build a rounded, fingertip-shaped GelSight capable of producing depth maps \cite{romero2020soft, romero2022soft}. However, as with other GelSight sensors, great care was taken to ensure a uniform distribution of all colored lights using light piping methods. Users desiring changes to the shapes or sizes of these sensors may therefore require further optical experiments and design alterations. Do and Kennedy also developed a semi-hemispherical sensor that utilized an autoencoder network to generate high resolution point clouds of the gel’s deformation with a mean error of 0.28mm \cite{do2022densetact}.

\subsection{All-Around Camera-Based Tactile Sensors}
The ability for a tactile fingertip-like sensor to sense multiple contacts from different directions, just like the human fingertip, can greatly help a robot speed up exploration time, sense unforeseen collisions, or even perform complex in-hand manipulation tasks that might not have been possible with a flatter sensor \cite{padmanabha2020omnitact, romero2020soft}. Though one-sided, curved sensors might be sufficient for parallel-jaw grippers, robot hands utilizing a greater number of fingers with higher DOFs may benefit from utilizing data provided by all-around sensors \cite{ahn2019robel, chen2022system, romero2020soft}.  
Recently, a wide range of camera-based, all-around sensors have been introduced. The OmniTact sensor \cite{padmanabha2020omnitact} has a finger-like form factor (cylindrical base with a semi-hemispherical top) and five strategically-placed cameras embedded in the painted gel elastomer. The sensor was able to estimate the angle of contact with objects and was successfully used in an insertion task \cite{padmanabha2020omnitact}. The GelTip sensor uses a geometry similar to \cite{padmanabha2020omnitact} but with a rigid shell supporting the gel. It uses a single camera housed at the sensor’s base and structured light to observe contact regions \cite{gomes2020geltip}. Sun et al. used the information from a structured-light illumination scheme with a neural network to estimate force measurements and contact localizations within sub-millimeter accuracy \cite{sun2022soft}. 

We build off the techniques used in the works mentioned above to introduce an all-around, curved GelSight sensor capable of producing high resolution depth reconstructions in a compact form factor. In addition, we introduce a generalized cross-illumination scheme that can be implemented in sensors of different shapes and sizes, allowing for the use of photometric stereo techniques to obtain object contact geometries.

\section{Design and Fabrication}

\subsection{Design Criteria} 
As the complexity and variety of dexterous robotic manipulation tasks have grown, a wide range of end effectors have been introduced to help accomplish these tasks \cite{ahn2019robel}. Since robotic grippers can vary in size, shape, and DOFs available, fitting them with camera-based tactile sensors that utilize specific lighting and optical strategies can be difficult and time consuming. The illumination strategy of our omnidirectional GelSight sensor aims to minimize these disadvantages. Specifically, we design our sensor to have 1) a small, compact tactile sensor and housing assembly that can modularly be fitted to different grippers 2) a 3D, rounded sensor shape with omnidirectional, high resolution tactile sensing capabilities, and 3) a generalizable cross-lighting illumination system that can be implemented in different sensor geometries. This section describes the design choices and fabrication methods used to construct a tactile sensor capable of meeting these criteria. 
\begin{figure}[!h]
    \centering
    \includegraphics[scale = 0.47]{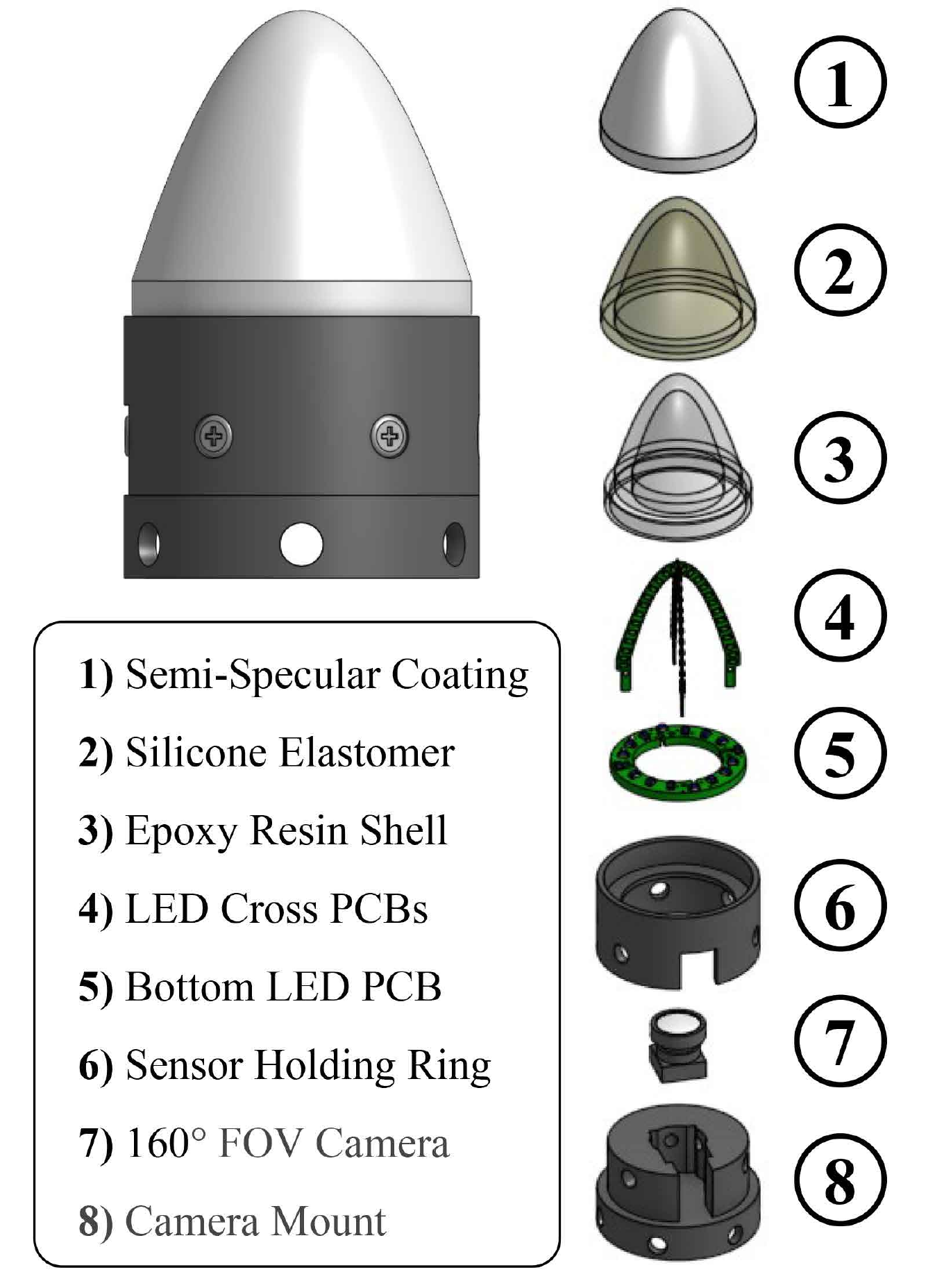}
    \caption{\footnotesize{\textbf{(Top Left)} 3D model of fully assembled sensor in housing structure. \textbf{(Right)} Exploded view of sensor.}}
    \label{fig:sensorassembly}
\end{figure}

\subsection{Sensor Illumination Strategy}

The lighting strategy of previous GelSight and GelSlim sensors used photometric stereo techniques to estimate the surface gradients generated when an object is pressed into the sensor’s soft, elastomer skin \cite{johnson2009retrographic, yuan2017gelsight, taylor2022gelslim, wang2021gelsight}. The RGB pixel intensities of the image produced by the sensor would then be mapped to the surface gradients with a look-up-table \cite{romero2020soft}. Once mapped, the horizontal and vertical gradients found could be integrated using the Fast Poisson Solver to produce a 3D height map of the sensor’s deformed elastomer skin.

\begin{figure}[!h]
    \centering
    \includegraphics[width=0.75\linewidth]{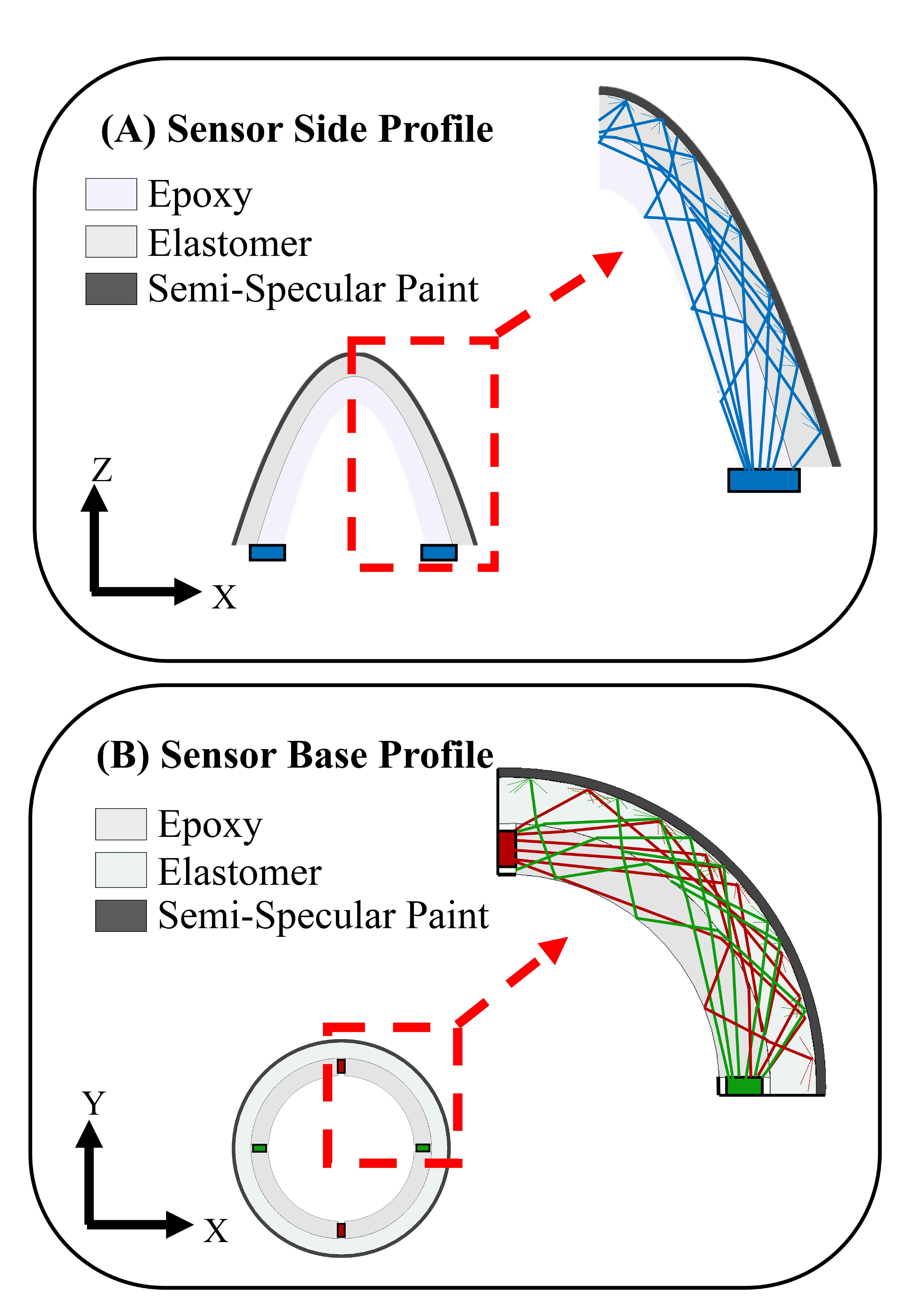}
    \caption{\footnotesize{Sensor illumination strategy implemented in the sensor to achieve photometric stereo. The top diagram shows a side view of the sensor with the blue LEDs shining from the base of the sensor. The light travels through the epoxy and elastomer, and is reflected by the elastomer's semi-specular coating. This provides the vertical blue color directionality. The bottom diagram shows a top-down, cross-sectional view of the sensor, where the red and green LEDs are implemented in an orthogonal, cross shape. The red and green lights travel through the epoxy and elastomer, and are partially reflected by the semi-specular paint. For simplicity, all reflections of the ray tracing are not shown.}}
    \label{fig:illuminationExample}
\end{figure}\label{crossidea}

Although the photometric stereo approach has worked well for GelSight sensors with flatter or partially curved surfaces \cite{romero2020soft, romero2022soft}, the omnidirectional, curved nature of our sensor requires a novel illumination strategy in order to use these techniques. In addition, we aim to develop a lighting system that can easily be customized to different sensor sizes and shapes without the need to use optical simulation software or repeated manufacturing attempts to adequately illuminate the different geometries. 

We transform the 2D red, green, and blue illumination pattern shown in \cite{wang2021gelsight} to a 3D-cross structure that uses ultra-thin LEDs mounted onto thin PCBs that are fitted together, as shown in Figure \ref{fig:sensorassembly} Part 4. Each PCB in the internal crossing structure is mounted on both sides with either red or green LEDs, providing two colored lights in orthogonal directions to each other in the XZ and YZ planes. To minimize the occlusions of the sensing surface caused by the PCBs and LEDs, the double-sided boards are manufactured to have 0.2mm thickness and 0.25mm thick SMD LEDs (1608 package). Reflow soldering is used to further minimize the cross-structure's thickness. The bottom LED mounted with blue LEDs (2012 package) provides the third illumination color in the XY plane. The blue LED ring also houses the VCC and GND through-hole connections for both the red and green LED boards, the appropriate resistors that control the current for each of the RGB circuit legs, and the connections that connect the sensor to the 5V and GND of the Raspberry Pi. This was done to help keep the electrical components contained within the sensor body and reduce wiring, helping make the sensor more modular for different gripper applications. The crossing-PCB structure is encapsulated in an epoxy resin shell and covered in a soft, gel elastomer covered in a semi-specular coating (described in Sections \ref{EpoxyShell} and \ref{GelCoating}). By simply altering the shapes and sizes of the PCBs, the 3D-cross structure can be used to produce different omnidirectional tactile sensors.

\subsection{Rigid Internal Skeleton Fabrication}\label{EpoxyShell}
The ultra-thin PCB boards used in the cross-illumination structure are prone to bending and breaking even under small deflections. To prevent the boards from breaking and to inhibit the LEDs from losing their orthogonal-plane configuration when the sensor is used, the boards must be encapsulated in a rigid resin shell matching the desired board outline. A two-part, gravity mold is designed to cast both the resin shell and gel elastomer. The mold negatives are printed on the Form Labs Form 2 Stereolithography printer, and Smooth-On Mold Star 20T is used to cast the silicone molds.   

To assemble the molding structure, the red and green LED boards are aligned onto the top silicone molding piece, and the bottom 5V and GND wirepads are threaded through the cut-outs made in the base of the silicone mold. 
We chose Smooth-On’s EpoxAcast 690 epoxy resin since it produces a bubble-free, durable shell when fully cured. The epoxy is mixed, degassed, and injected into the top inlet holes of the mold. It is left to cure at room temperature for at least 24 hours.
\begin{figure}[!h]
    \centering
    \includegraphics[width=0.9\linewidth]{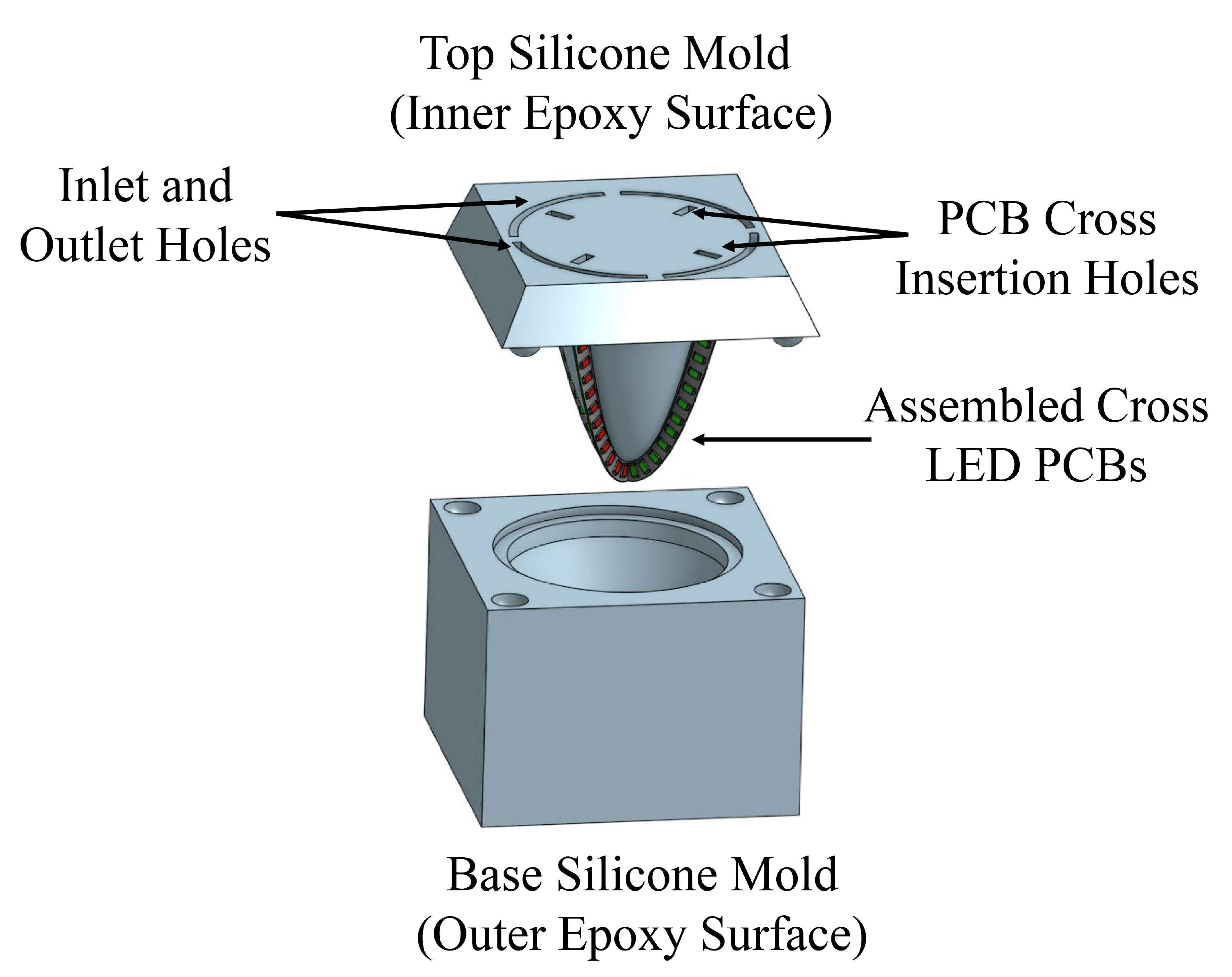}
    \vspace{-3mm}
    \caption{\footnotesize{Molding strategy used to encapsulate the flimsy, ultra-thin PCBs in a hard epoxy shell to prevent breaking and ensure the orthogonal direction of the red and green LEDs is maintained.}}
    \label{fig:moldingExample}
\end{figure}

\subsection{Painted Gel Elastomer Fabrication}
\label{GelCoating}

The molds for casting the soft, outer elastomer are made using the same method as described in Section \ref{EpoxyShell}. Additionally, the top mold covering used when casting the resin shell is re-used to ensure that the resin shell and elastomer coating remain aligned on all axes throughout this process. Since we aim to ultimately use these sensors to add tactile feedback to robotic grippers that will be repeatedly interacting with objects, the durability after prolonged use must be considered. To promote the adhesion that reduces the chance of delamination between the internal resin shell and the soft elastomer, the surface of the epoxy is brushed with DOWSIL P5200 Adhesion Promoter. 

We choose an opaque semi-specular aluminum coating for our sensor, as it is more sensitive to changes in the sensor’s surface normals and provides less diffusion of light when compared to other Lambertian, matte coatings used in previous, flatter tactile sensors \cite{yuan2017gelsight, wang2021gelsight, taylor2022gelslim}. The silicone molds are prepared by first spraying the molds with a thin layer of Mann Ease Release 200, followed by airbrushing the inside surface of the base mold with an opaque coating 
of 1 : 1 : 0.25 : 3 of the Smooth-On Psycho Paint Part A, Smooth-On Psycho Paint Part B, aluminum flakes, and NOVOCS solvent. We find that by painting the molds and filling them with the uncured PDMS,  the 
silicone and paint can cure together, making it less likely for the outer paint to wrinkle, delaminate, or be scratched off over the sensor’s lifetime. A ratio of 1:18 (Activator : Base) of optically clear Silicones Inc. XP-565 is mixed, degassed, and poured into the painted base mold. The resin skeleton is then pushed into the base mold and left to cure at room temperature overnight. 
\subsection{Assembly and Housing} \label{Assembly and Housing}
Once the PDMS has cured, the sensor is demolded, and the blue LED ring is press-fit into the inner rim of the sensor. 
The exposed wires from the epoxy-encapsulated LED boards and Raspberry Pi 5V/GND wires are soldered into the PCB ring.

The housing for the sensor is 3D printed with Onyx filament on the Markforged Onyx One printer. The bottom rim of the sensor is covered in a thin layer of Gorilla Clear Epoxy Glue and press fit into the sensor holder ring (see Figure \ref{fig:sensorassembly} Part 6). Our sensor design uses the 160\degree{}  Frank-S01-V1.0 Raspberry Pi Spy Camera since it is able to see the entire desired sensing surface. The camera is focused and press-fit into the camera holder piece. M2 screws are used to attach the camera housing to the sensor housing. An outlet hole in the housing allows the LEDs to be connected to the 5V/GND pins and the CSI cable for the camera to be connected to the Raspberry Pi. 

\section{Sensor Calibration for Depth Map Generation} \label{calibrationAlgo}

\begin{figure*}[!h]
    \centering
    \includegraphics[width=0.98\linewidth]{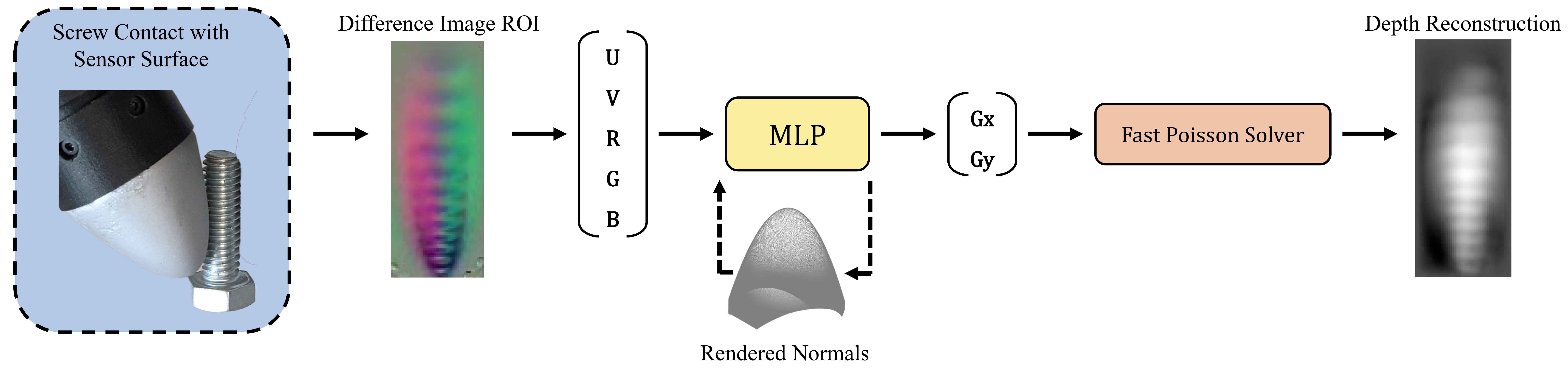}
    \caption{\footnotesize{Depth reconstruction pipeline described in Section \ref{calibrationAlgo}. The (u,v) image coordinates, along with the RGB values of the normalized difference image of the contact area are input into an MLP to estimate the deformed surface gradients. The gradients are then input into a Fast Poisson Solver to generate depth reconstructions. The MLP is trained using the rendered normals produced by probing the sensor with a 4mm sphere in known world coordinates.}}
    \label{fig:pipeline}
\end{figure*}

Previous flat-surfaced, Lambertian-coated GelSight and related sensors are calibrated by repeatedly pressing a ball with known radius (around 2mm to 3mm) over the surface of the sensor \cite{wang2021gelsight, yuan2017gelsight}. After obtaining the pixel to millimeter ratio in the sensor’s image, the horizontal and vertical gradients (Gx and Gy) of the pixels in contact with the pressed sphere can be calculated. A lookup table linearly mapping the pixel’s color information and location to its gradient can then be obtained. However, because of our sensor’s curved shape and semi-specular coating, the same methods could not be used. Romero et al. and Do et al. were able to reconstruct the surface of their curved, non-Lambertian sensors using a nonlinear function to map the pixel colors and locations to their respective gradients using deep learning networks. We implement methods similar to \cite{romero2022soft, romero2020soft, wang2021gelsight, do2022densetact} to obtain the 3D reconstructions of the deformations on the sensor’s surface.

Our calibration and reconstruction procedure consists of the following steps: 1) Probing the entire surface of the sensor with a 4mm sphere at known world coordinates to obtain the pixel locations (u,v) and color intensities (RGB values), 2) Using the sensor geometry and the known probing locations to generate the approximate surface normals at these points, and 3) Finding a continuous, nonlinear mapping function using a multilayer-perceptron (MLP) network to generate the gradients that can be used with the Fast Poisson Solver.  
\subsection{Sensor Calibration}
Since we aim to introduce a sensor utilizing a cross-illumination structure capable of working for a variety of shapes and sizes, our method relies on the accuracy of the camera matrices, calibration probing locations, and 2D-image to 3D-world-coordinate correspondences to accurately generate the gradients of the probed contact region. Although it is possible to use ray-casting methods and conversions to spherical/cylindrical coordinate systems like in \cite{do2022densetact, gomes2020geltip}, we aim to present a generalized procedure that can be implemented on a variety of all-around sensor shapes. 

1) \textbf{CNC Calibration}: To satisfy the requirements, we use the CNC probing setup used in \cite{romero2020soft} since the high positional accuracy of the machine will probe the sensor at the desired coordinates. A mesh model of the sensor’s surface in the real-world (whose offsets have been corrected due to 3D printing tolerances in the sensor’s casing) is sampled to choose N probing coordinates. The assembled sensor is rigidly mounted to the base of the CNC, and a 3D-printed probe with a 4mm calibration ball at its end travels to each coordinate location and presses the ball into the surface of the sensor. The camera located inside the sensor captures the calibration ball’s impression at each of the N locations. Our results use N = 10,000 probing coordinates, which takes about 12 hours to collect. 

2) \textbf{2D-image coordinates to 3D-world coordinates Correspondence}: In order to use an off-the-shelf rendering software to generate the gradients of the sensor, we find the camera’s intrinsic matrices, distortion coefficients, and pose in the world coordinate frame. Before the sensor housing is attached to the camera housing (as described in Section \ref{Assembly and Housing}), the focused 160\degree{} FOV camera is calibrated 
using OpenCV’s fisheye lens camera calibration procedure \cite{bradski2000opencv}. Using the intrinsic matrix and distortion coefficients, all of the internal sensor images collected during the CNC probing are undistorted. Next, 100 randomly sampled CNC probed points of the sensor’s surface are used to find the camera’s pose in the world coordinate frame. Although we have an approximate estimate of the camera’s physical location in space from the sensor assembly’s CAD model, the camera pose with respect to the optical center is needed for accurate normals generation. For each of the undistorted sampled images, the minimum point of the indented sphere (the point closest to the camera) is chosen in the image coordinates, and the corresponding minimum point is chosen from a point cloud that simulates the ball indenting into the sensor’s surface. The camera extrinsic matrix is found using the RANSAC PnP computation, and the camera pose in the world coordinate frame is calculated. 

3)	\textbf{Collecting Gradients and Contact ROI}: Using the camera matrices and sensor/calibration ball CAD models, rendering software is used to estimate the intersection of the two models. Here, we assume the ball is rigid and able to fully penetrate the sensor’s surface since the elastomer itself is only 1.75mm thick. The simulation serves two purposes: 1) To generate the internal Gx and Gy normals of the sensor’s surface when probed with the calibration ball corresponding to each pixel image point and 2) To generate a binary mask of the ball’s contact region for collecting the desired pixel coordinates and intensity values from the probed images. Past GelSight works that used flat and curved surfaces \cite{wang2021gelsight, romero2020soft} were able to use a Hough Circle Transform to correctly isolate the contact regions of the probed ball during calibration. However, since the distance from the camera to the sensor surface can vastly vary due to the height of the sensor, and the curvature or sloping of the sensor’s surface can distort the circle, tuning the parameters to find the circles proved to be unreliable. We therefore opted to generate the contact region mask using the rendering software, as it reliably provided accurate masks for the different sensor shapes tested. 

\subsection{Normal Estimation with MLP}
Finally, we use the mask to collect the coordinate locations (u,v) and pixel intensities (RGB) of the areas in contact with the calibration ball, along with their respective gradient values (Gx, Gy). 
The continuous lookup table is implemented with a multi-layer perceptron (MLP) network trained on the contact region. As in \cite{wang2021gelsight}, we balance the data by adding in around 25\% non-contact region data. 

\section{Results}

\subsection{Different Sensor Geometries}
\begin{figure}[!h]
    \centering
    \includegraphics[width=\linewidth]{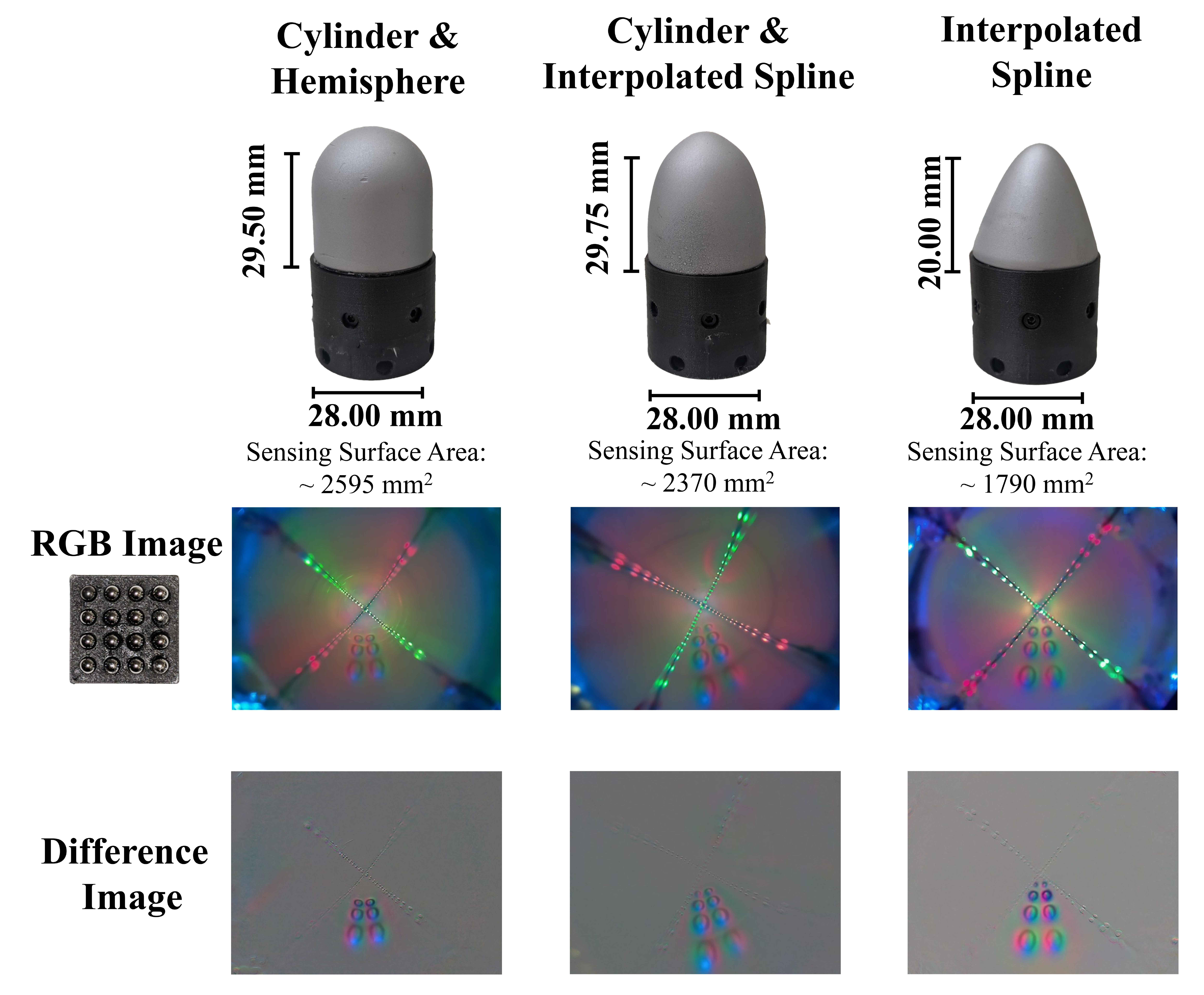}
    \caption{\footnotesize{Examples of different sensor shapes manufactured using the cross-illumination strategy described in Section \ref{crossidea}. The second row shows example raw RGB images collected when an array of 3mm steel balls is pressed into the surface of the sensor. The third row shows example difference images, highlighting the directionality of the different LED colors. 
    }}
    \label{fig:shapeExamples}
\end{figure}
The cross-illumination strategy is tested in the three geometries shown in Figure \ref{fig:shapeExamples}. With the exception of altering the molds to produce the desired shapes and changing the PCB board outlines, the overall manufacturing process of different sensors remains the same. Additionally, slight differences in resistor values and camera color gains are necessary to obtain high quality color information since the heights of the sensors differed (Note: between sensors of the same size and shape, these values remained constant). The raw RGB and difference images (see Figure \ref{fig:shapeExamples}) show how the expected color directionality necessary to achieve photometric stereo can uniformly be seen throughout most of the sensor’s body.



We do note, however, that new designs may be somewhat constrained to the focal length of the camera and diameter of the camera lens. Because of the vertically-oriented nature of our all-around sensor, the camera may not be properly focused for the entire body of the sensor; building a sensor whose tip is at a large distance from the camera may impact the resolution of the tactile data. Therefore, the camera’s optical characteristics should be considered when customizing the sensors. Additionally, the types of environments and tasks that the robot will be performing should be taken into account. For example, tasks that require the robot to utilize tactile information coming from the tip of the sensor would benefit more from the cylindrical + hemispherical configuration, since the surface area of the sensor greatly decreases towards the ends of the fingers in conical geometries. At the top of the conical sensors, a large proportion of pixels providing data at the top of the sensor are also occluded by the cross-LED structure, lowering the visibility in this region. In tasks that require digging, singulation, or rotational actions, sensing on the sides of the fingers would be more advantageous, making the conical shaped fingertips a better choice for the application.

\subsection{Depth Map Reconstruction}

\begin{figure}[!h]
    \centering
    \includegraphics[width=\linewidth]{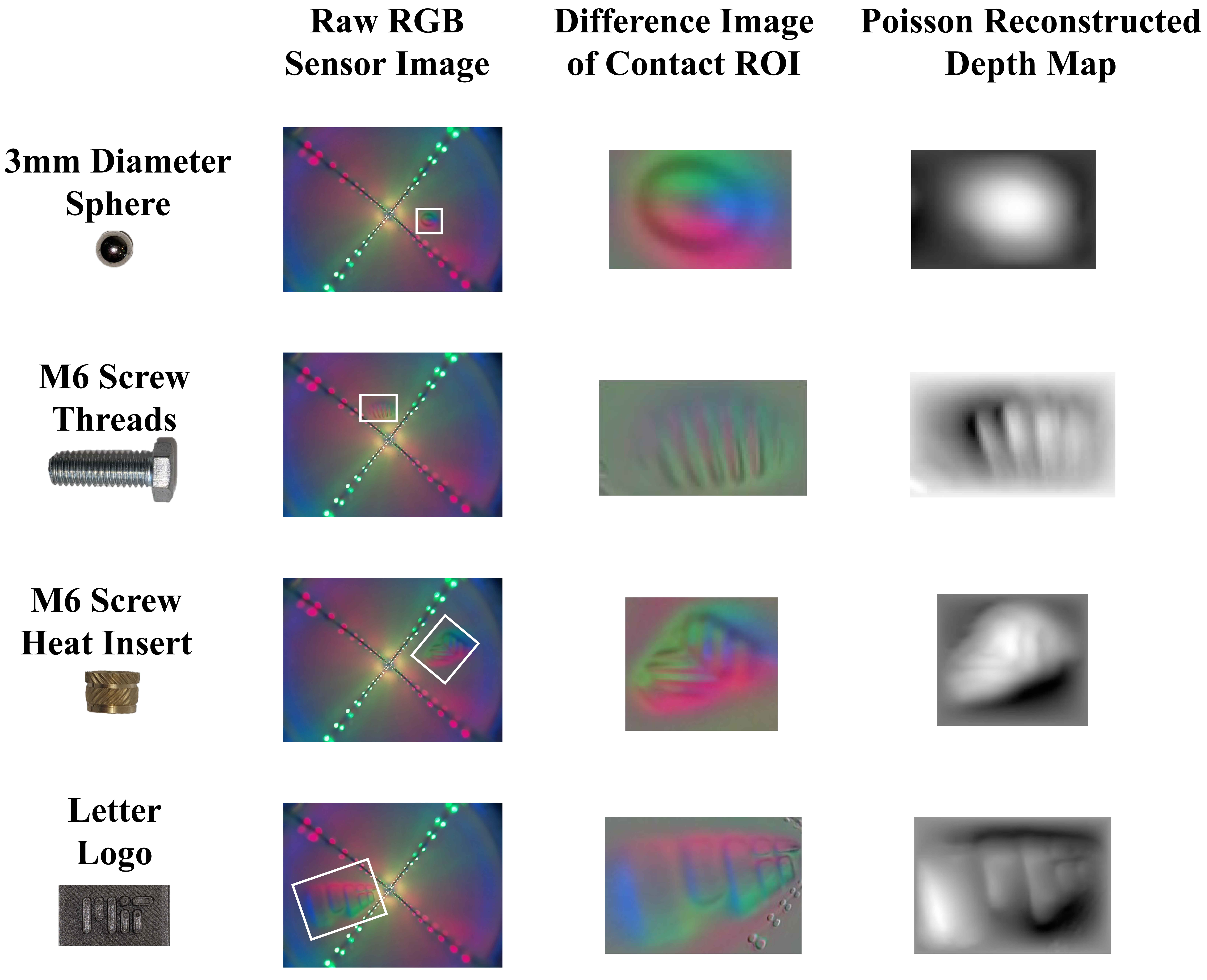}
    \caption{\footnotesize{Examples of tactile images collected when a variety of objects are pressed into the sensor's surface. The second column shows the raw RGB image collected from the sensor. The third column shows the difference image of the contact region of interest. Finally, the last column shows the approximate depth map reconstruction after the calibration procedure described in Section \ref{calibrationAlgo}.}}
    \label{fig:reconstruction}
\end{figure}

Because of the sensor’s semi-specular aluminum coating and the highly curved surface of the sensor, we implement a continuous look-up-table using an MLP. When an unknown object is pressed into the sensor, the network is used to estimate the Gx and Gy values of the contact patch. These values are fed into the Fast Poisson Solver to obtain an approximate 3D depth map of the deformed contact region. Figure \ref{fig:reconstruction} shows the depth maps produced from the algorithm for a variety of objects being pressed into different quadrants and heights on the spline shaped sensor. Though the entire pipeline described in Section \ref{calibrationAlgo} can be implemented for the other shapes mentioned, we show the reconstructions of the spline-shaped sensor, since its shape is unique to past omni-directional sensors \cite{padmanabha2020omnitact, gomes2020geltip, do2022densetact} and was predicted to be difficult to obtain reconstructions from due to its shape and more drastic non-uniformity in lighting when compared to the other shapes tested. 

Overall, the general shape and some details of the contact areas of most small objects pressed into the sensor can be seen, such as the threads of the M6 screw, the ridges on the M6 heat insert, and even the letters in the logo block. However, the accuracy of the reconstructed images have shown to share some relationship with the location of the contact area and the geometry of the object itself. In general, more rounded objects whose contact location is not locally linear tend to produce depth maps with less accuracy. Additionally, areas closer to the top of the sensor, where most of the directional colored light comes from only the blue color channel, have a harder time estimating the Gx gradients (due to the lack of red and green color), causing somewhat warped reconstructions.

\subsection{Occlusions and Un-Uniform Lighting}
\begin{figure}[!ht]
    \centering
    \includegraphics[width=0.5\textwidth]{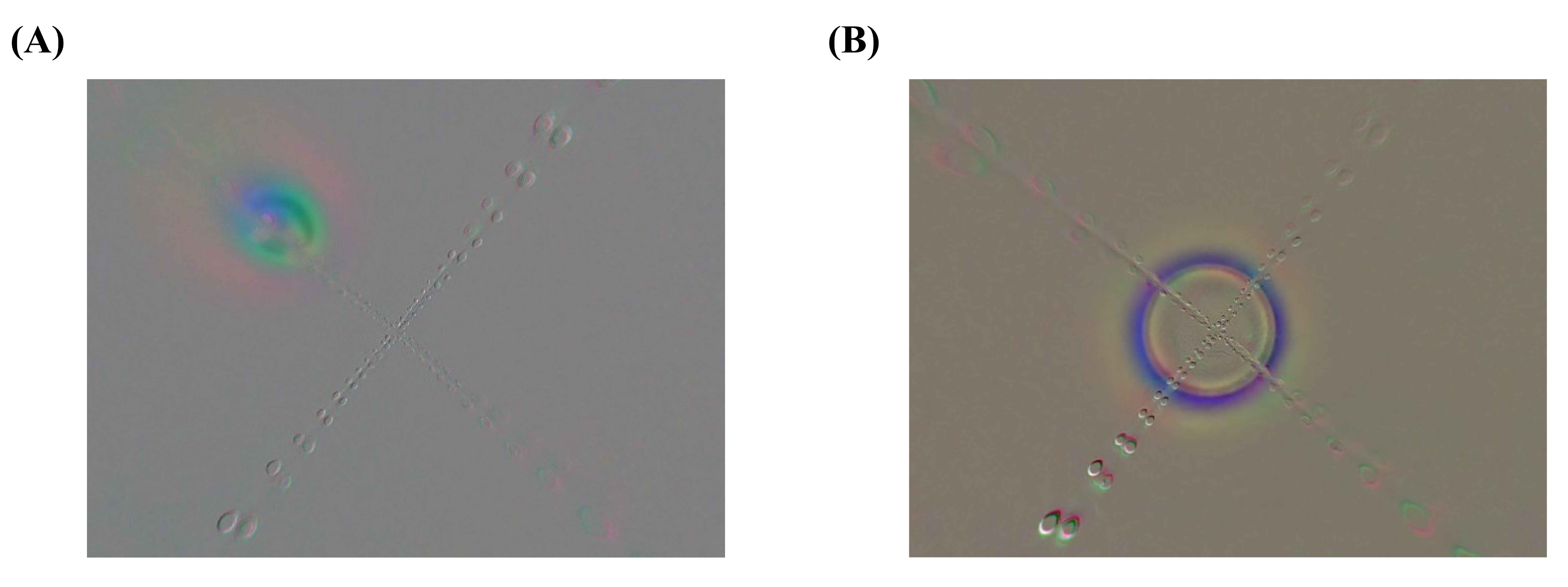}
    \caption{\footnotesize{Examples of lighting and occlusion occurrences in the sensor. \textbf{(A)} Example of the pixel occlusions that can occur due to the cross-illumination strategy. The line of gray blocking out the center of the 3mm calibration ball helps visualize the data lost due to the PCBs and LEDs. \textbf{(B)} Difference image collected when a ring is pressed onto the top of the sensor. The impression only exhibits directional light from the blue color channel, causing issues in predicting the Gx gradients for depth reconstruction. }}
    \label{fig:drawbacks}
\end{figure}

One drawback of using the cross-illumination strategy is the inevitable visual occlusions that occur due to the illuminating LED fins in the epoxy shell. Approximately 10\% of the viable pixels in the undistorted image are blocked by the  LEDs. Even though the thinnest LEDs and boards were used to produce the desired directional lighting scheme, their obstructions to seeing the entire surface of the sensor are noted. Image A in Figure \ref{fig:drawbacks} shows an example of the obstruction that occurs when a 3mm steel sphere is pressed into the area of the surface adjacent to the lights. As the contact areas get farther away from the camera, the occlusions do slightly worsen. In cases where the object in contact is larger than the width of the fins, we can assume that the depth of the object is similar to areas fully visible to the sensor, allowing us to interpolate depths along these specific pixels, similar to the methods used in \cite{wang2021gelsight}

A possible source of error in the reconstructed depth images is the lack of directional lighting occurring from all three color channels towards the top of the sensor. Throughout the base and mid-section of the sensor, the independent directional lighting from the red, green, and blue color channels can be distinctly observed (see Figure \ref{fig:reconstruction}). However, as the light sources in the red and green channel grow farther away and the camera’s sensor becomes less sensitive to those wavelengths, the blue color channel shinning directly upwards from the base of the sensor floods the contact area, making it harder for the network to predict the correct gradients. An example of this can be seen in Image B of Figure \ref{fig:drawbacks}, where a 3D-printed ring is placed around the top of the sensor. This difference image shows no red or green illumination.

\section{CONCLUSION}
In this work, we present an all-around tactile sensing fingertip that implements a novel cross-illumination system to achieve photometric stereo constraints and produce high resolution depth reconstructions. In addition, we show that this lighting scheme can be implemented in different sizes and geometries without requiring intensive optical simulation and experimentation previously required to manufacture curved sensors. The sensor is calibrated by pressing a small ball into known locations on the sensor’s surface, and the gradients of the contact regions are rendered and input into a neural network to learn the nonlinear function mapping the image pixel’s location and color information to the gradients. The surface normals are then fed into the Fast Poisson Solver to produce depth maps of the object contact regions of the surface. However, there are some pitfalls to the sensor, such as occlusions caused by LEDs located in the sensor’s shell and the lack of perfectly uniform lighting throughout the sensor. 

Future work includes reconfiguring the cross structure to minimize occlusions and further increase uniform lighting, studying methods of interpolating the depth over the occluded surfaces, adding force and other sensing modalities to the sensors, and further assessing the advantages of omnidirectional sensors in manipulation tasks. We also refer readers to \cite{pai2023tactofind}, for continued works employing GelSight360 sensors on a 3-fingered, 9 DOF robotic gripper \cite{ahn2019robel} to successfully complete an object retrieval task using only tactile feedback. By introducing a generalizable design strategy to easily manufacture and calibrate compact, all-around tactile sensors in a variety of shapes and sizes, roboticists can easily augment their systems with tactile feedback,  further improving a robot’s ability to successfully complete assembly, exploration, or home-care tasks.





\section*{ACKNOWLEDGMENT}
This material is based upon work supported by the National Science Foundation
Graduate Research Fellowship under Grant No. [1745302], Toyota Research Institute (TRI), the Office of Naval Research (ONR), the SINTEF GentleMAN program, and the Amazon Science Hub. 
The authors would also like to thank Branden Romero for his helpful discussions on the manufacturing and calibration of curved tactile sensors, Sandra Q. Liu for her advice on sensor fabrication, and Pulkit Agrawal, Abhishek Gupta, and Tao Chen for their suggestions on sensor improvements for use in manipulation tasks.

\bibliographystyle{IEEEtran}
\bibliography{gelsight360}

\begin{thebibliography}{10}
\providecommand{\url}[1]{#1}
\csname url@rmstyle\endcsname
\providecommand{\newblock}{\relax}
\providecommand{\bibinfo}[2]{#2}
\providecommand\BIBentrySTDinterwordspacing{\spaceskip=0pt\relax}
\providecommand\BIBentryALTinterwordstretchfactor{4}
\providecommand\BIBentryALTinterwordspacing{\spaceskip=\fontdimen2\font plus
\BIBentryALTinterwordstretchfactor\fontdimen3\font minus
  \fontdimen4\font\relax}
\providecommand\BIBforeignlanguage[2]{{%
\expandafter\ifx\csname l@#1\endcsname\relax
\typeout{** WARNING: IEEEtran.bst: No hyphenation pattern has been}%
\typeout{** loaded for the language `#1'. Using the pattern for}%
\typeout{** the default language instead.}%
\else
\language=\csname l@#1\endcsname
\fi
#2}}

\bibitem{wang2021gelsight}
S.~Wang, Y.~She, B.~Romero, and E.~Adelson, ``Gelsight wedge: Measuring
  high-resolution 3d contact geometry with a compact robot finger,'' in
  \emph{2021 IEEE International Conference on Robotics and Automation
  (ICRA)}.\hskip 1em plus 0.5em minus 0.4em\relax IEEE, 2021, pp. 6468--6475.

\bibitem{alspach2019soft}
A.~Alspach, K.~Hashimoto, N.~Kuppuswamy, and R.~Tedrake, ``Soft-bubble: A
  highly compliant dense geometry tactile sensor for robot manipulation,'' in
  \emph{2019 2nd IEEE International Conference on Soft Robotics
  (RoboSoft)}.\hskip 1em plus 0.5em minus 0.4em\relax IEEE, 2019, pp. 597--604.

\bibitem{she2021cable}
Y.~She, S.~Wang, S.~Dong, N.~Sunil, A.~Rodriguez, and E.~Adelson, ``Cable
  manipulation with a tactile-reactive gripper,'' \emph{The International
  Journal of Robotics Research}, vol.~40, no. 12-14, pp. 1385--1401, 2021.

\bibitem{liu2022gelsight}
S.~Q. Liu and E.~H. Adelson, ``Gelsight fin ray: Incorporating tactile sensing
  into a soft compliant robotic gripper,'' in \emph{2022 IEEE 5th International
  Conference on Soft Robotics (RoboSoft)}.\hskip 1em plus 0.5em minus
  0.4em\relax IEEE, 2022, pp. 925--931.

\bibitem{yuan2016estimating}
W.~Yuan, M.~A. Srinivasan, and E.~H. Adelson, ``Estimating object hardness with
  a gelsight touch sensor,'' in \emph{2016 IEEE/RSJ International Conference on
  Intelligent Robots and Systems (IROS)}.\hskip 1em plus 0.5em minus
  0.4em\relax IEEE, 2016, pp. 208--215.

\bibitem{kappassov2015tactile}
Z.~Kappassov, J.-A. Corrales, and V.~Perdereau, ``Tactile sensing in dexterous
  robot hands,'' \emph{Robotics and Autonomous Systems}, vol.~74, pp. 195--220,
  2015.

\bibitem{zainuddin2015resistive}
N.~Zainuddin, N.~F. Anuar, A.~L. Mansur, N.~I.~M. Fauzi, W.~F. Hanim, and S.~H.
  Herman, ``Resistive-based sensor system for prosthetic fingers application,''
  \emph{Procedia Computer Science}, vol.~76, pp. 323--329, 2015.

\bibitem{muhlbacher2015responsive}
S.~M{\"u}hlbacher-Karrer, A.~Gaschler, and H.~Zangl, ``Responsive
  fingers—capacitive sensing during object manipulation,'' in \emph{2015
  IEEE/RSJ International Conference on Intelligent Robots and Systems
  (IROS)}.\hskip 1em plus 0.5em minus 0.4em\relax IEEE, 2015, pp. 4394--4401.

\bibitem{heyneman2012biologically}
B.~Heyneman and M.~R. Cutkosky, ``Biologically inspired tactile classification
  of object-hand and object-world interactions,'' in \emph{2012 IEEE
  International Conference on Robotics and Biomimetics (ROBIO)}.\hskip 1em plus
  0.5em minus 0.4em\relax IEEE, 2012, pp. 167--173.

\bibitem{sundaram2019learning}
S.~Sundaram, P.~Kellnhofer, Y.~Li, J.-Y. Zhu, A.~Torralba, and W.~Matusik,
  ``Learning the signatures of the human grasp using a scalable tactile
  glove,'' \emph{Nature}, vol. 569, no. 7758, pp. 698--702, 2019.

\bibitem{fishel2012sensing}
J.~A. Fishel and G.~E. Loeb, ``Sensing tactile microvibrations with the
  biotac—comparison with human sensitivity,'' in \emph{2012 4th IEEE RAS \&
  EMBS international conference on biomedical robotics and biomechatronics
  (BioRob)}.\hskip 1em plus 0.5em minus 0.4em\relax IEEE, 2012, pp. 1122--1127.

\bibitem{ntagios2020robotic}
M.~Ntagios, H.~Nassar, A.~Pullanchiyodan, W.~T. Navaraj, and R.~Dahiya,
  ``Robotic hands with intrinsic tactile sensing via 3d printed soft pressure
  sensors,'' \emph{Advanced Intelligent Systems}, vol.~2, no.~6, p. 1900080,
  2020.

\bibitem{zhang2022finger}
J.~Zhang, H.~Yao, J.~Mo, S.~Chen, Y.~Xie, S.~Ma, R.~Chen, T.~Luo, W.~Ling,
  L.~Qin, \emph{et~al.}, ``Finger-inspired rigid-soft hybrid tactile sensor
  with superior sensitivity at high frequency,'' \emph{Nature communications},
  vol.~13, no.~1, pp. 1--9, 2022.

\bibitem{yong2022soft}
S.~Yong, J.~Chapman, and K.~Aw, ``Soft and flexible large-strain piezoresistive
  sensors: On implementing proprioception, object classification and curvature
  estimation systems in adaptive, human-like robot hands,'' \emph{Sensors and
  Actuators A: Physical}, vol. 341, p. 113609, 2022.

\bibitem{tang2019design}
Z.~Tang, Z.~Wang, J.~Lu, and G.~Ma, ``Design of robot finger based on flexible
  tactile sensor,'' \emph{International Journal of Advanced Robotic Systems},
  vol.~16, no.~5, p. 1729881419879853, 2019.

\bibitem{bhirangi2021reskin}
R.~Bhirangi, T.~Hellebrekers, C.~Majidi, and A.~Gupta, ``Reskin: versatile,
  replaceable, lasting tactile skins,'' \emph{arXiv preprint arXiv:2111.00071},
  2021.

\bibitem{ward2018tactip}
B.~Ward-Cherrier, N.~Pestell, L.~Cramphorn, B.~Winstone, M.~E. Giannaccini,
  J.~Rossiter, and N.~F. Lepora, ``The tactip family: Soft optical tactile
  sensors with 3d-printed biomimetic morphologies,'' \emph{Soft robotics},
  vol.~5, no.~2, pp. 216--227, 2018.

\bibitem{lambeta2020digit}
M.~Lambeta, P.-W. Chou, S.~Tian, B.~Yang, B.~Maloon, V.~R. Most, D.~Stroud,
  R.~Santos, A.~Byagowi, G.~Kammerer, \emph{et~al.}, ``Digit: A novel design
  for a low-cost compact high-resolution tactile sensor with application to
  in-hand manipulation,'' \emph{IEEE Robotics and Automation Letters}, vol.~5,
  no.~3, pp. 3838--3845, 2020.

\bibitem{shimonomura2019tactile}
K.~Shimonomura, ``Tactile image sensors employing camera: A review,''
  \emph{Sensors}, vol.~19, no.~18, p. 3933, 2019.

\bibitem{yamaguchi2016combining}
A.~Yamaguchi and C.~G. Atkeson, ``Combining finger vision and optical tactile
  sensing: Reducing and handling errors while cutting vegetables,'' in
  \emph{2016 IEEE-RAS 16th International Conference on Humanoid Robots
  (Humanoids)}.\hskip 1em plus 0.5em minus 0.4em\relax IEEE, 2016, pp.
  1045--1051.

\bibitem{li2014localization}
R.~Li, R.~Platt, W.~Yuan, A.~ten Pas, N.~Roscup, M.~A. Srinivasan, and
  E.~Adelson, ``Localization and manipulation of small parts using gelsight
  tactile sensing,'' in \emph{2014 IEEE/RSJ International Conference on
  Intelligent Robots and Systems}.\hskip 1em plus 0.5em minus 0.4em\relax IEEE,
  2014, pp. 3988--3993.

\bibitem{sun2022soft}
H.~Sun, K.~J. Kuchenbecker, and G.~Martius, ``A soft thumb-sized vision-based
  sensor with accurate all-round force perception,'' \emph{Nature Machine
  Intelligence}, vol.~4, no.~2, pp. 135--145, 2022.

\bibitem{taylor2022gelslim}
I.~H. Taylor, S.~Dong, and A.~Rodriguez, ``Gelslim 3.0: High-resolution
  measurement of shape, force and slip in a compact tactile-sensing finger,''
  in \emph{2022 International Conference on Robotics and Automation
  (ICRA)}.\hskip 1em plus 0.5em minus 0.4em\relax IEEE, 2022, pp.
  10\,781--10\,787.

\bibitem{yuan2017gelsight}
W.~Yuan, S.~Dong, and E.~H. Adelson, ``Gelsight: High-resolution robot tactile
  sensors for estimating geometry and force,'' \emph{Sensors}, vol.~17, no.~12,
  p. 2762, 2017.

\bibitem{yuan2015measurement}
W.~Yuan, R.~Li, M.~A. Srinivasan, and E.~H. Adelson, ``Measurement of shear and
  slip with a gelsight tactile sensor,'' in \emph{2015 IEEE International
  Conference on Robotics and Automation (ICRA)}.\hskip 1em plus 0.5em minus
  0.4em\relax IEEE, 2015, pp. 304--311.

\bibitem{li2018slip}
J.~Li, S.~Dong, and E.~Adelson, ``Slip detection with combined tactile and
  visual information,'' in \emph{2018 IEEE International Conference on Robotics
  and Automation (ICRA)}.\hskip 1em plus 0.5em minus 0.4em\relax IEEE, 2018,
  pp. 7772--7777.

\bibitem{do2022densetact}
W.~K. Do and M.~Kennedy, ``Densetact: Optical tactile sensor for dense shape
  reconstruction,'' in \emph{2022 International Conference on Robotics and
  Automation (ICRA)}.\hskip 1em plus 0.5em minus 0.4em\relax IEEE, 2022, pp.
  6188--6194.

\bibitem{yin2022multimodal}
J.~Yin, G.~M. Campbell, J.~Pikul, and M.~Yim, ``Multimodal proximity and
  visuotactile sensing with a selectively transmissive soft membrane,'' in
  \emph{2022 IEEE 5th International Conference on Soft Robotics
  (RoboSoft)}.\hskip 1em plus 0.5em minus 0.4em\relax IEEE, 2022, pp. 802--808.

\bibitem{johnson2009retrographic}
M.~K. Johnson and E.~H. Adelson, ``Retrographic sensing for the measurement of
  surface texture and shape,'' in \emph{2009 IEEE Conference on Computer Vision
  and Pattern Recognition}.\hskip 1em plus 0.5em minus 0.4em\relax IEEE, 2009,
  pp. 1070--1077.

\bibitem{li2013sensing}
R.~Li and E.~H. Adelson, ``Sensing and recognizing surface textures using a
  gelsight sensor,'' in \emph{Proceedings of the IEEE Conference on Computer
  Vision and Pattern Recognition}, 2013, pp. 1241--1247.

\bibitem{yuan2018active}
W.~Yuan, Y.~Mo, S.~Wang, and E.~H. Adelson, ``Active clothing material
  perception using tactile sensing and deep learning,'' in \emph{2018 IEEE
  International Conference on Robotics and Automation (ICRA)}.\hskip 1em plus
  0.5em minus 0.4em\relax IEEE, 2018, pp. 4842--4849.

\bibitem{bauza2019tactile}
M.~Bauza, O.~Canal, and A.~Rodriguez, ``Tactile mapping and localization from
  high-resolution tactile imprints,'' in \emph{2019 International Conference on
  Robotics and Automation (ICRA)}.\hskip 1em plus 0.5em minus 0.4em\relax IEEE,
  2019, pp. 3811--3817.

\bibitem{huang2022understanding}
H.-J. Huang, X.~Guo, and W.~Yuan, ``Understanding dynamic tactile sensing for
  liquid property estimation,'' \emph{arXiv preprint arXiv:2205.08771}, 2022.

\bibitem{romero2020soft}
B.~Romero, F.~Veiga, and E.~Adelson, ``Soft, round, high resolution tactile
  fingertip sensors for dexterous robotic manipulation,'' in \emph{2020 IEEE
  International Conference on Robotics and Automation (ICRA)}.\hskip 1em plus
  0.5em minus 0.4em\relax IEEE, 2020, pp. 4796--4802.

\bibitem{patel2020digger}
R.~Patel, R.~Ouyang, B.~Romero, and E.~Adelson, ``Digger finger: Gelsight
  tactile sensor for object identification inside granular media,'' in
  \emph{International Symposium on Experimental Robotics}.\hskip 1em plus 0.5em
  minus 0.4em\relax Springer, 2020, pp. 105--115.

\bibitem{romero2022soft}
B.~R. Romero, ``Soft, round, high resolution tactile fingertip sensors for
  dexterous robotic manipulation,'' 2022, p.~54.

\bibitem{padmanabha2020omnitact}
A.~Padmanabha, F.~Ebert, S.~Tian, R.~Calandra, C.~Finn, and S.~Levine,
  ``Omnitact: A multi-directional high-resolution touch sensor,'' in \emph{2020
  IEEE International Conference on Robotics and Automation (ICRA)}.\hskip 1em
  plus 0.5em minus 0.4em\relax IEEE, 2020, pp. 618--624.

\bibitem{ahn2019robel}
M.~Ahn, H.~Zhu, K.~Hartikainen, H.~Ponte, A.~Gupta, S.~Levine, and V.~Kumar,
  ``Robel: Robotics benchmarks for learning with low-cost robots. arxiv
  e-prints, page,'' \emph{arXiv preprint arXiv:1909.11639}, 2019.

\bibitem{chen2022system}
T.~Chen, J.~Xu, and P.~Agrawal, ``A system for general in-hand object
  re-orientation,'' in \emph{Conference on Robot Learning}.\hskip 1em plus
  0.5em minus 0.4em\relax PMLR, 2022, pp. 297--307.

\bibitem{gomes2020geltip}
D.~F. Gomes, Z.~Lin, and S.~Luo, ``Geltip: A finger-shaped optical tactile
  sensor for robotic manipulation,'' in \emph{2020 IEEE/RSJ International
  Conference on Intelligent Robots and Systems (IROS)}.\hskip 1em plus 0.5em
  minus 0.4em\relax IEEE, 2020, pp. 9903--9909.

\bibitem{bradski2000opencv}
G.~Bradski, ``{The OpenCV Library},'' \emph{Dr. Dobb's Journal of Software
  Tools}, 2000.

\bibitem{pai2023tactofind}
S.~Pai, T.~Chen, M.~Tippur, E.~Adelson, A.~Gupta, and P.~Agrawal, ``Tactofind:
  A tactile only system for object retrieval,'' in \emph{2023 IEEE
  International Conference on Robotics and Automation (ICRA)}, 2023.

\end{thebibliography}

\end{document}